\documentclass[10pt,twocolumn,letterpaper]{article}

\usepackage{cvpr}
\usepackage{times}
\usepackage{epsfig}
\usepackage{graphicx}
\usepackage{amsmath}
\usepackage{amssymb}
\usepackage{algorithmic}
\usepackage{algorithm}
\usepackage{memhfixc}
\usepackage{amstext}
\usepackage{fixltx2e}
\usepackage{paralist}
\usepackage{placeins}
\usepackage{fancyhdr}

\usepackage{cite}
\usepackage{booktabs}	

\DeclareMathOperator*{\argmin}{argmin}

\usepackage[pagebackref=true,breaklinks=true,letterpaper=true,colorlinks,bookmarks=false]{hyperref}

\cvprfinalcopy 

\def\cvprPaperID{1483} 

\ifcvprfinal\fi

\begin{document}



\title{\emph{Features in Concert:} Discriminative Feature Selection meets
       \\ Unsupervised Clustering
}

\author{
\begin{tabular}{ccc}
Marius Leordeanu$^{1,2}$ & Alexandra Radu$^{2}$ & Rahul Sukthankar$^{3}$ \\
\end{tabular}\\
\\
$^1$Institute of Mathematics of the Romanian Academy\\
$^2$Faculty of Automatic Control and Computer Science, University Politehnica of Bucharest\\
$^3$Google Research
}


\maketitle

\begin{abstract}
Feature selection is an
essential problem in computer vision, important
for category learning and recognition.
Along with the rapid development of a wide variety of visual features
and classifiers, there is a growing need for efficient
feature selection and combination methods, to construct
powerful classifiers for more complex and higher-level recognition tasks.
We propose an algorithm that efficiently discovers sparse,
compact representations of input features or classifiers, from a vast sea of
candidates, with important optimality properties, low computational
cost and excellent accuracy in practice.
Different from boosting, we start with a discriminant linear classification formulation
that encourages sparse solutions. Then we obtain an equivalent
unsupervised clustering problem that
jointly discovers ensembles of diverse features. They are independently valuable
but even more powerful when united in a cluster of classifiers.
We evaluate our method on the task of large-scale recognition in video and show that it
significantly outperforms classical selection approaches, such as AdaBoost and greedy
forward-backward selection, and powerful classifiers such as SVMs, in speed of training
and performance, especially in the case of limited training data.
\end{abstract}

\section{Introduction}

\begin{figure}
\begin{center}
\includegraphics[scale = 0.3, angle = 0, viewport = 0 0 900 680, clip]{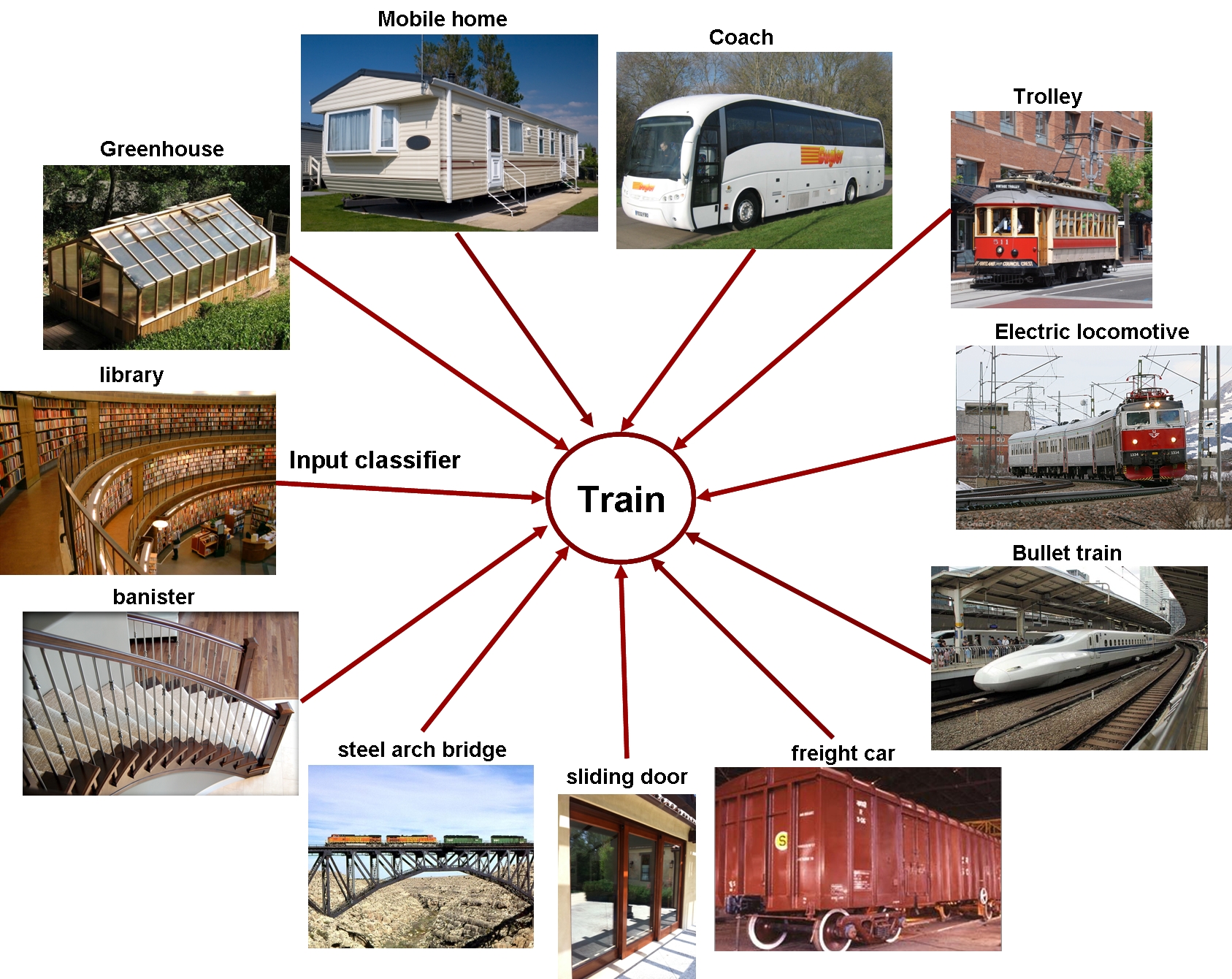}
\caption{Context in the mind: What classes can \emph{trigger} the idea of
a ``train''?
Many classes have similar appearance but are semantically unrelated;
others are semantically close but visually dissimilar.
We argue that consistently co-firing classifiers, either based on spatial
and temporal context or similar appearance, can be powerful in collaboration
and robust to outliers, overfitting and missing features.
Here, we show the classifiers that are consistently
selected by our method, from very limited training data, as
providing valuable input to the class ``train''.
}
\label{fig:teaser_train}
\end{center}
\end{figure}

The design of efficient ensembles of classifiers
has proved very useful over decades of computer vision
and machine learning research~\cite{vasconcelos2003feature,dietterich2000ensemble},
with applications to virtually all classification tasks addressed,
ranging from detection of specific
types of objects, such as human faces~\cite{ViJo04},
to more general mid- and higher-level category recognition problems.
There is a growing sea of potential visual features and classifiers,
whether manually designed or automatically learned. They have the potential
to participate in building powerful classifiers on new classification problems.
Often classes are \emph{triggered} by only a few key input features
(Fig.~\ref{fig:teaser_train}). Objects and object categories can be
identified by the presence of certain discriminative keypoints~\cite{key:lowe,key:mutch_lowe},
or discriminative collections of weaker features~\cite{ViJo04,key:leordeanu_cvpr07},
and higher-level human actions and more complex video activities can be categorized by
certain key frames, poses or relations between body parts~\cite{ellis_rahul_ijcv2012,cuntoor_accv2006,zanfir2013moving}.
The development of efficient feature discovery and combination methods
for learning new concepts could have a strong impact in real
world applications.

Feature selection is known to be NP-hard~\cite{guyon2003introduction,ng1998feature},
so finding the optimal solution to the combinatorial search
is prohibitive.
Thus, previous work has focused on greedy methods, such as sequential
search~\cite{pudil1994floating} and boosting~\cite{freund1995decision}
or heuristic approaches, such as genetic algorithms~\cite{siedlecki1989note}.
We approach feature selection from a different direction, that of discriminant linear
classification~\cite{DuHa73}, with a novel constraint on the solution and the features.
We put an upper bound on the solution weights and further require it to
be an affine combination of soft-categorical features,
which have to be also positively correlated
with the positive class. Our constraints
lead to a convex formulation with some important theoretical guarantees that
strongly favor sparse optimal solutions with equal non-zero weights.
This automatically becomes a feature selection mechanism, such that most features with zero weights can be ignored
while the remaining few are averaged to become a strong group of classifiers
with a single united voice.

Consider Fig.~\ref{fig:teaser_train}: here we use image-level CNN classifiers~\cite{jia2014caffe}, pre-trained on ImageNet, to recognize trains in video frames from YouTube-Objects dataset~\cite{prest2012learning}.
Our method builds an ensemble from a pool of $6000$ classifiers ($1000$ ImageNet classifiers $\times$ $6$ image regions) that are potentially relevant to the concept.
Since each classifier corresponds to one ImageNet concept, we directly visualize some of the classifiers
(shown as sample images from corresponding classes)
that are consistently selected by our method
over $30$ trials on different small sets of $8$ video shots,
each with just $10$ evenly spaced frames.
We observe that the classes chosen may seem semantically different from
train (e.g. library, greenhouse, steel bridge),
but they are definitely related to the concept,
either through appearance (e.g. library, greenhouse), through
context (steel bridge), or both (sliding door).

\section{Scientific Context}
Decades of research in machine learning show that an ensemble can be significantly
stronger than an individual classifier in isolation~\cite{dietterich2000ensemble,hansen1990neural},
especially when the individual classifiers are diverse and make mistakes
on different regions of the input space.
There are many methods for ensemble learning that have been studied over the years~\cite{maclin2011popular,dietterich2000ensemble},
with three main approaches: \emph{bagging}~\cite{breiman1996bagging}, \emph{boosting}~\cite{freund1995decision}
and \emph{decision trees ensembles}~\cite{kwok2013multiple,criminisi2012decision}.

Bagging blindly samples from the training set to learn a different classifier for each sampled
set, then takes the average response over all classifiers as the final answer. While this approach
avoids overfitting, it does not explore deeper structure in the data and, in practice, the same classifier type
is used for each random training subset. Different from bagging we select small subsets of relevant features
over the whole training set. Our feature pool contains diverse and potentially strong classifiers (Fig.~\ref{fig:diverse_features}), either created from scratch or reused from
pre-trained libraries (Sec.~\ref{sec:feature_types}).


Boosting is a popular technique that in general outperforms bagging,
as it searches for relevant features from a
vast pool of candidates. It adds features one by one, in an efficient greedy fashion,
to reduce the expected exponential loss.
The sequential addition of features
puts much more weight on the initial ones selected.
If too much weight is given to the first features (when they are strong classifiers by themselves),
boosting is less expected to form powerful classifier ensembles that help each other as a group,
as the initial features selected will dominate. Thus, boosting works best with weak features,
and has difficulty with more powerful ones, such as SVMs~\cite{li2008adaboost}. Our method
is well suited for combining strong classifiers,
which together form an even stronger group. They are discovered as clusters of co-firing
classifiers that are independent given the class, but united on separating the positive class versus the rest.
The balanced collaboration between classifiers encourages similar weights for each input feature. In turn, equal averaging
leads to classifier independence given the class (Sec.~\ref{sec:theory}).

\begin{figure}
\begin{center}
\includegraphics[scale = 0.37, angle = 0, viewport = 0 0 500 520, clip]{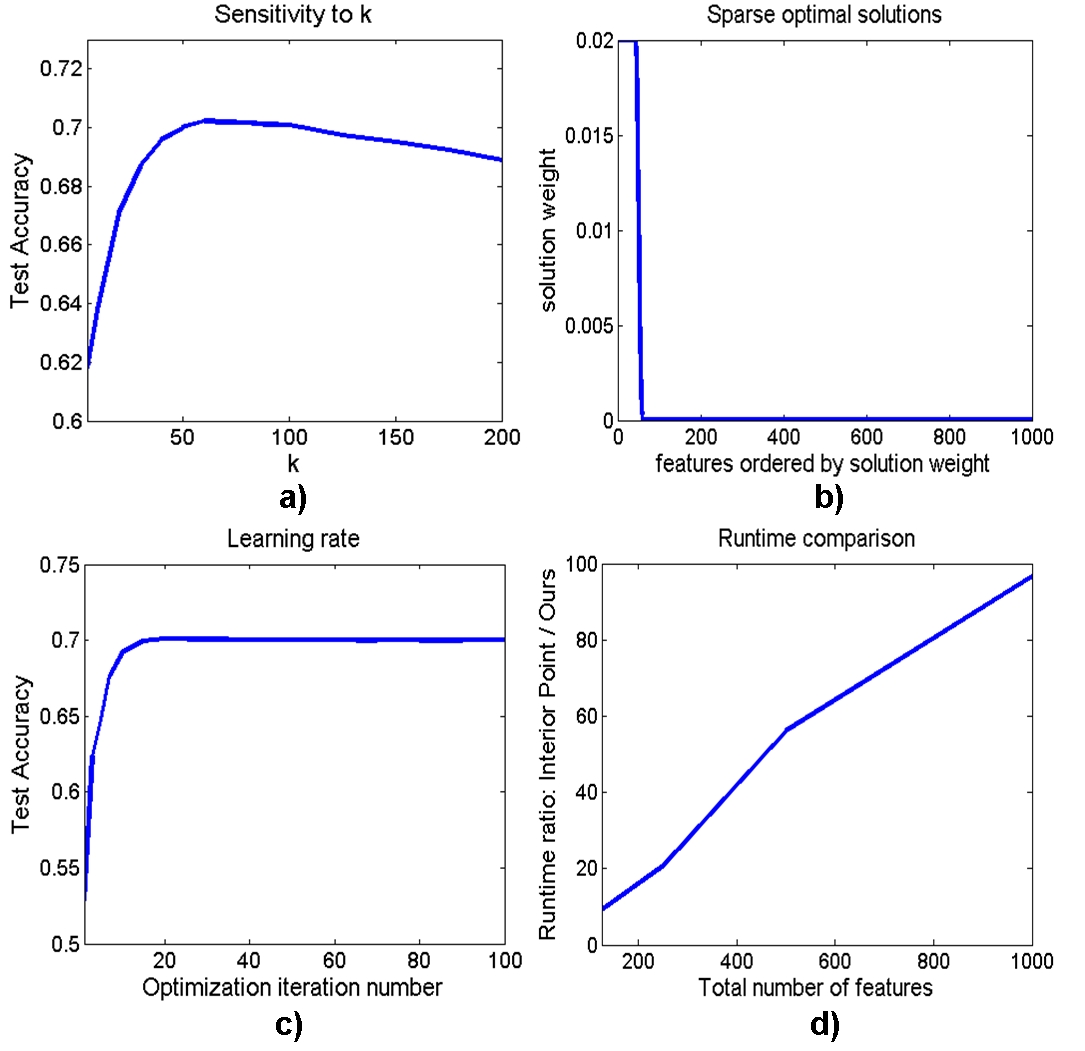}
\caption{Optimization and sensitivity analysis:
a) Sensitivity to $k$.
Performance improves as features are added, is stable
around the peak $k=60$ and falls $k>100$ as useful features
are exhausted.
b) Features ordered by weight for $k=50$ confirming that our method selects
nearly equal weights up to the chosen $k$.
c) Our method converges to a solution in $10$--$20$ iterations.
d) Runtime of interior point method divided by ours,
both in Matlab and with $100$ max iterations. All results are averages
over $100$ random experiments.
}
\label{fig:optimization}
\end{center}
\end{figure}

Our method is also related to averaging decision trees.
One of the main differences is that we do not average
\emph{all} of the classifiers: we identify the few most important ones and average over them.
Averaging over a judicious
set rather than blind averaging over the pool
makes a significant difference (Fig.~\ref{fig:optimization}a).
There is also work~\cite{schulter2013alternating}
on combining decision forests with ideas from boosting, in order to obtain a weighted average of trees that better fits the training data.
Rather than consuming a significant amount of training data to fit optimal
weights, our method focuses on finding subsets of features that will work
well with known similar weights. By averaging strong subsets of diverse classifiers
we obtain excellent accuracy and generalization, even from limited training data.

We are not the first to see a connection between clustering and feature selection.
Some consider the inverse task: feature selection for unsupervised
clustering~\cite{yang2011l2,law2004simultaneous}. Others
propose efficient selection of features through
diversity~\cite{vasconcelos2003feature}. However, we are the first to
formulate supervised learning as an equivalent unsupervised clustering task.

In Section \ref{sec:feature_types}, we describe in more detail
how we create novel powerful features by
naturally clustering
the training data over neighborhoods in descriptor space (CIFAR features),
contiguous temporal regions in time (Youtube-Parts features) and spatial neighborhoods over
different image windows/regions of presence (Youtube-Parts and ImageNet features). They provide
intermediate lower level classifiers for the higher level problem of category understanding,
in the presence of significant variations in scale, poses and viewpoints, intr-class variations, and
large background clutter. These intermediate features could be seen as as building blocks in
a hierarchical and potentially recursive recognition system, validating some of the ideas in~\cite{leordeanu2014thoughts}.

The connection to hierarchical approaches based on Deep Nets~\cite{hinton_deep_learning_2006,hinton_RBM_2010}
is interesting, both from a feature creation and re-usability perspective, as well as from
the viewpoint of building multi-layered hierarchical classifiers. The relation to other hierarchical
approaches is also beneficial, given the many successful hierarchical approaches
in computer vision, from the classifier cascades used for face detection~\cite{ViJo04},
the Part-Based Model and Latent SVMs~\cite{felzenszwalb_ObjDetect_pami2010} applied to general object
category detection, Conditional Random Fields~\cite{quattoni_hiddenCRFs_2007}, classification trees and random forests,
probabilistic Bayesian networks, directed acyclic graphs (DAGs)~\cite{jensen_nielsen_2007},
hierarchical hidden Markov models (HHMMs)~\cite{HMMM_1998} and methods using feature matching with second-order
or hierarchical spatial constraints~\cite{key:leordeanu_cvpr07,Conte_GraphMatching_2004,lazebnik2006beyond}.

\paragraph{Main Contributions:}The contributions of our novel approach to
learning discriminative sparse classifier averages are summarized below:

\begin{enumerate}
\item A novel approach to linear classification that is equivalent
to unsupervised learning defined as a convex quadratic program, with efficient
optimization. The global solution
is sparse with equal weights effectively leading to a feature selection
procedure. This is important since feature selection is known
to be NP-hard.

\item An efficient clustering method that is one to two orders of magnitude
faster in practice than interior point convex optimization, based on
recent work on the IPFP algorithm~\cite{key:leordeanu_IPFP} and the Frank-Wolfe
method~\cite{key:FW}.

\item Compared to more sophisticated methods, such as Ada\-Boost and SVM, our algorithm exhibits better generalization with more modest computational and storage costs. Our training time is quadratic in the number of available features but constant in the number of training samples.

\item Efficient ways of automatically constructing powerful intermediate features as classifiers learned
from various datasets (Section \ref{sec:feature_types}). This transfers knowledge from different image classification tasks to a new problem
of recognition in video and provides the ability to re-use resources by transforming previously learned classifiers
into input features to novel learning tasks. While learning auto-encoders~\cite{hinton_deep_learning_2006,rifai2011contractive})
also effectively uses anonymous classifiers as input features to higher level interpretation
layers, we provide a way to use apparently unrelated classifiers, learned from different data,
as black boxes. Our linear discriminant approach to feature selection becomes an effective procedure
of learning one layer at a time and further validates some of our proposals in~\cite{leordeanu2014thoughts}
\end{enumerate}

\section{Problem Formulation}
\label{sec:formulation}

We address the classical case of binary classification, with the one vs.\ all
strategy being applied to the multi-class scenario as well.
Given a set of $N$ training samples, with each $i$-th sample expressed
as a vector $\mathbf{f}_i$ of $n$ possible features with values between $0$ and $1$,
we want to find the weight vector $\mathbf{w}$, with non-negative elements
and L1-norm $1$, such that $\mathbf{w}^T\mathbf{f}_i \approx p_1$ when the $i$-th sample is
from class $1$ and $\mathbf{w}^T\mathbf{f}_i \approx p_0$ otherwise.
As $p_0$ and $p_1$ represent the expected feature average output for negative and positive samples, respectively, then $0 \le p_0 \le p_1 \le 1$.
We require the input features $\mathbf{f}_i$ to be positively correlated
with class $1$; when they are not we simply \emph{flip} their output, by setting
$f_i(j) \gets 1 - f_i(j)$. Traditionally, $p_0 = 0$ and $p_1 = 1$, but we used $p_0 = 0$ and $p_1 = 0.5$,
with slightly improved performance, as averages over positives are expected to be less than $1$.

In order to limit the impact of each individual feature
we restrict the elements of $\mathbf{w}$ to be between $0$ and $1/k$, and sum up to $1$.
Our formulation is similar to linear classification with the added constraints that the input features
themselves could represent other classifiers and the linear separator $\mathbf{w}$ acts as an affine combination
of their outputs, to produce a weighted feature average $\mathbf{w}^T\mathbf{f}_i \in [0,1]$.
In Section \ref{sec:theory} we show that the value of $k$ has a direct role on the sparsity of
the solution and the number of features that have strong weights, a fact validated by our experiments.

Given the $N \times n$ feature data matrix $\mathbf{F}$ and ground
truth vector $\mathbf{t}$, the learning problem becomes finding
$\mathbf{w^*}$ that minimizes the sum of squares error
$J(\mathbf{w}) = \|\mathbf{Fw} - \mathbf{t}\|^2$,
under the constraints on $\mathbf{w}$. We obtain the convex problem:
\begin{eqnarray}
\label{eq:learning}
\mathbf{w^*} & = & \argmin_w J(\mathbf{w}) = \argmin_w \|\mathbf{Fw} - \mathbf{t}\|^2 \\
    & = & \argmin_w \mathbf{w^\top (F^\top F) w} - 2\mathbf{(Ft)}^\top \mathbf{w} + \mathbf{t}^\top \mathbf{t} \nonumber \\
    & & s.t. \sum_i w_i =1 \;, \; w_i \in [0,1/k]. \nonumber
\end{eqnarray}
\noindent Since $\mathbf{t}$ is the ground truth, the last term is constant.
After dropping it, we note that the supervised learning task is a special case
of clustering with pairwise and unary terms, as defined in
~\cite{key:bulo_nips09,key:latecki_nips10,leordeanu2012efficient}.
Note that our formulation can be easily changed into a concave maximization
problem by changing the signs of the terms.
Since the algorithm of~\cite{leordeanu2012efficient} works with both positive and negative
terms, we adapt their efficient optimization scheme that achieves near-optimal solutions
in only $10-20$ iterations.

The connection to clustering is interesting and makes sense. Feature selection
can be interpreted as a clustering problem: we seek a group of features
that are individually relevant, but not redundant with respect to each other~--- an observation consistent with earlier
research in machine learning (e.g.,~\cite{dietterich2000ensemble})
and neuroscience (e.g.,~\cite{rolls2010noisy}).
This idea is also related to the recent work on discovering discriminative groups of
HOG filters~\cite{ahmed2014knowing}, but different from that and other previous work,
in that ours transforms the supervised learning task into an equivalent unsupervised clustering
problem. To get a better intuition let us examine in more detail the two terms
of the objective, the quadratic one $\mathbf{w^\top (F^\top F) w}$ and the
linear term $- 2\mathbf{(Ft)}^\top \mathbf{w}$. If we assume that feature outputs have similar means
and standard deviations over training samples
(a fact that could be obtained by appropriate normalization), then
minimizing the linear term boils down
to giving more weight to features that are more strongly correlated with the ground truth.
This is expected, since they are the ones that are best for classification by themselves. On the other hand, the
matrix $\mathbf{F^\top F}$ contains the dot-products between pairs of feature responses
over the training set. Then, minimizing $\mathbf{w^\top (F^\top F) w}$
should find groups of features
that are as uncorrelated as possible.
The value of $1/k$ limits the weight put on any single input classifier and requires the final solution
to have nonzero weights for at least $k$ features. In Section~\ref{sec:theory}
we present analysis that the solution
preferred is sparse, very often having exactly $k$ features with uniform weights of value exactly $1/k$.

\section{Theoretical Analysis}
\label{sec:theory}

The optimization problem is convex and can be globally solved
in polynomial time. We adapted the integer projected
fixed point method from~\cite{leordeanu2012efficient} to the case of unary and pairwise
terms, which is very efficient in practice (Fig.~\ref{fig:optimization}c).
The optimization procedure is iterative
and approximates at each step the original error function
with a linear, first-order Taylor approximation that can be solved immediately.
That step is followed by a \emph{line search} with rapid closed-form solution, and the process is repeated
until convergence. Please see~\cite{leordeanu2012efficient,key:leordeanu_IPFP} for more
details. In practice, after only $10$--$20$
iterations we are very close to the optimum, but we used $100$ iterations
in all our experiments. The theoretical guarantees at the optimum
prove that Problem~\ref{eq:learning} prefers sparse solution with
equal weights, also confirmed in practice (Fig.~\ref{fig:optimization}b).

\noindent \textbf{Proposition 1:}
Let $\mathbf{d(w)} = 2\mathbf{F^\top Fw - F^\top t}$ be the gradient
of $J(\mathbf{w})$. The partial derivatives $d(\mathbf{w})_i$ corresponding
to those elements $w^*_i$ of the global optimum of Problem~\ref{eq:learning}
with non-sparse, real values in $(0,1/k)$ must be equal to each other.

\noindent \textbf{Proof:}
The global optimum of Problem~\ref{eq:learning}
satisfies the Karush-Kuhn-Tucker (KKT) necessary optimality conditions.
The Lagrangian function of (\ref{eq:learning}) is:

\begin{eqnarray}
L(\mathbf{w}, \lambda, \mu , \beta) &=& J(\mathbf{w}) - \lambda (\sum w_i - 1) + \nonumber \\
& & \sum \mu_i w_i + \sum \beta_i (1/k - w_i),
\label{eq:lagrangian}
\end{eqnarray}
From the KKT conditions at a point $\mathbf{w^*}$ we have:
\[
\begin{array}{l}
\mathbf{d(w^*)} - \lambda + \mu_i - \beta_i = 0,\\
\sum_{i=1}^{n} \mu_i w^*_i = 0,\\
\sum_{i=1}^{n} \beta_i (1/k - w^*_i) = 0.\\
\end{array}
\]
Here $\mathbf{w^*}$ and the Lagrange multipliers have non-negative elements,
so if $w_i > 0 \Rightarrow \mu_i = 0$
and $w_i < 1/k \Rightarrow \beta_i = 0 $. Then there must exist
a constant $\lambda$ such that we have:
\[d(\mathbf{w^*}) =  \left\{
\begin{array}{ll}
 \leq \lambda, & w^*_i = 0, \\
 = \lambda,    & w^*_i \in (0, 1/k), \\
\geq \lambda, & w^*_i = 1/k. \\
\end{array}\right.
\]
This implies that all partial derivatives of $d(\mathbf{w^*})$ that are not in $[0,1/k]$
must be equal to some constant $\lambda$, therefore they must be equal to each other,
which concludes our proof.

From Proposition $1$ it follows that in the general case, when the
partial derivatives at the optimum point are unique, the elements of
the optimal $\mathbf{w^*}$ are either $0$ or $1/k$.
Since the sum over the elements of $\mathbf{w^*}$ is $1$, it is further implied that the
number of nonzero elements in $\mathbf{w^*}$ is often $k$.
Thus, our solution is not just a simple linear
separator (hyperplane), but also a sparse representation and a feature selection
procedure that effectively averages the selected $k$ or close to $k$ features.
To enable a better statistical interpretation of these sparse averages,
we consider the somewhat idealized case when all features
have equal means $(p_1, p_0)$ and equal standard deviations
$(\sigma_1, \sigma_0)$ over the positive and negative training sets, respectively.

\noindent \textbf{Proposition 2:}
If we assume that the input soft classifiers are independent
and better than random chance, the error rate converges towards $0$
as their number $n$ goes to infinity.

\noindent \textbf{Proof:}
Given a classification threshold $\theta$ for $\mathbf{w}^T\mathbf{f}_i$,
such that $p_0 < \theta < p_1$, then, as $n$ goes to infinity, the probability that a negative sample will have an
average response greater than $\theta$ (a false positive mistake) goes to $0$.
This follows from Chebyshev's inequality (or the Law of Large Numbers).
By a similar argument, the probability of a false negative also goes to zero as
$n$ goes to infinity.

\noindent \textbf{Proposition 3:}
The weighted average $\mathbf{w}^T\mathbf{f}_i$
with smallest variance over positives (and negatives, respectively)
has equal weights.

\noindent \textbf{Proof:} We consider the case when $\mathbf{f}_i$'s are features of positive
samples, the same argument being true for the negative ones.
We have: $\text{Var}(\sum_iw_if_i/\sum_iw_i) = \sum w_i^2/(\sum w_i)^2 \sigma^2_1$.
We find the minimum of  $\sum w_i^2/(\sum w_i)^2$ by setting its partial
derivatives to zero and obtain $w_j(\sum w_i)=\sum w_i^2, \forall j$. Therefore,
$w_i=w_j, \forall i,j$.

Equal weights minimize the
output variances over positives, and over negatives, separately (P3),
so they are most likely to minimize
the error rate, when the features are independent and follow
the equal means and variance assumptions above (P2).
This is important, since our method will certainly find
the set of features with equal weights (in general) that minimize
the convex error objective \ref{eq:learning} (P1).

\paragraph{Computational aspects:}
Compared to the general case
of arbitrary real weights for all possible features, the averaging solution
preferred by Problem~\ref{eq:learning}
requires considerably
less memory. The average of $k$ selected features out of $N$ possible
requires about $k\log_2 N$ bits, whereas having a real weight
for each possible feature requires $32N$ bits in floating point representation.
Sparse solutions are simpler
in terms of representation but have good accuracy and considerably smaller
computational cost (Fig.~\ref{fig:test_and_time}) than the more costly SVM and AdaBoost.
They seem to follow closer the Occam's Razor principle~\cite{blumer1987occam},
which would explain in part their good performance and generalization.
The computational cost of the optimization method we use is
$O(Sn^2)$~\cite{leordeanu2012efficient},
where $S$ is the number of iterations and $n$ is the number of features.
In our experiments we use $S=100$, even though $S=20$ would suffice.
The more general interior point method for convex optimization using Matlab's
$quadprog$ is polynomial, but considerably slower than ours, by a factor that increases
linearly with features pool size (see Fig.~\ref{fig:optimization}).
For $125$ features it is $9$ times slower, and for $1000$ features, about $100$ times slower.

\begin{figure}
\begin{center}
\includegraphics[scale = 0.27, angle = 0, viewport = 0 0 900 950, clip]{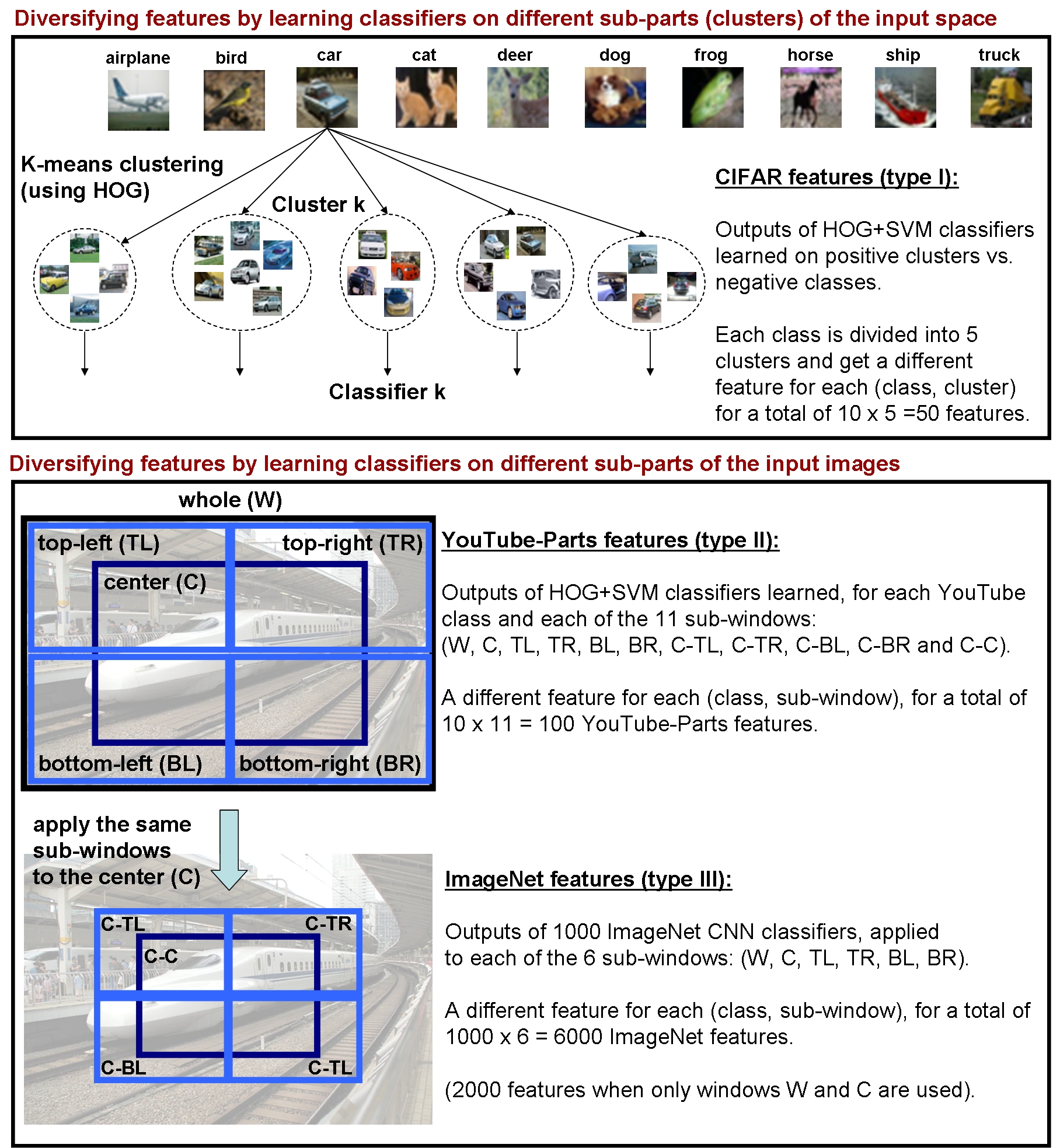}
\caption{We encourage feature diversity and independence by taking classifiers
trained on 3 datasets (CIFAR, YouTube-Objects and ImageNet) and by looking at different parts of the input space (Type~I) or different locations within the image (Types~II and~III). Experiments confirm the benefits of diversity. }
\label{fig:diverse_features}
\end{center}
\end{figure}

\section{Learning the Feature Pool}
\label{sec:feature_types}

We created a large pool of over $6000$ different features, computed and learned
from three different datasets: CIFAR~\cite{krizhevsky2009learning}, ImageNet~\cite{imagenet_cvpr09}
and a hold-out part of the YouTube-Objects training set. More details about creating our features
follow next and are also summarized in Fig.~\ref{fig:diverse_features}.

\paragraph{CIFAR features (type I):}
This dataset contains
$60000$ 32$\times$32 color images in $10$ classes
(airplane, automobile, bird, cat, deer, dog, frog, horse, ship, truck),
with $6000$ images per class.
There are 50000 training images and 10000 test images.
We randomly chose $2500$ images per class for creating our features. They are HOG+SVM
classifiers trained on data obtained by clustering images from each class into $5$ groups
using k-means applied to their HOG descriptors. Each classifier had to separate its own cluster
versus images from other classes. We hoped to obtain, for each class, diverse and relatively
independent classifiers, which respond to different parts of the input space that are naturally
clustered. Note that CIFAR categories coincide only partially ($7$ out of $10$ with the ones
from YouTube-Objects). The output of each of the $5\times10=50$ such classifiers becomes a different
input feature, which we compute on all training and test images from YouTube-Objects.

\paragraph{YouTube-parts features (type~II):}
We formed a separate dataset with $25000$ images from video, randomly selected from a subset
of YouTube-Objects Training videos, not used in subsequent learning and recognition experiments.
Features are outputs of linear SVM classifiers using HOG applied to the different parts of
each image. Each classifier is trained and applied to its own dedicated sub-window
as shown in Fig.~\ref{fig:diverse_features}. To speed up training and remove noise
we also applied PCA to the resulted HOG, and obtained descriptors of $46$ dimensions,
before passing them to SVM. For each of the $10$ classes, we have $11$ classifiers, one for
each sub-window, and get a total of $110$ type II features.
Experiments with a variety of
SVM kernels and settings showed that linear SVM with default parameters
for libsvm worked best, and we kept that fixed in all experiments.

\paragraph{ImageNet features (type~III):}
We considered the soft feature outputs (before soft max) of the pre-trained ImageNet CNN features using Caffe~\cite{jia2014caffe},
each of them over six different sub-windows: \emph{whole, center, top-left, top-right, bottom-left, bottom-right},
as presented in Fig.~\ref{fig:diverse_features}. There are $1000$ such outputs, one for each ImageNet category,
for each sub-window, for a total of $6000$ features. In some of our experiments, when specified, we
used only $2000$ ImageNet features, restricted to the \emph{whole} and \emph{center} windows.

\section{Experimental Analysis}
\label{sec:experiments}

We evaluate our method's ability to generalize and learn quickly from limited
data as well as transfer and combine knowledge from different datasets, containing
video or low- and medium-resolution images of many potentially
unrelated classes. We evaluate its performance in the context of recognition in video
and report recognition accuracy per frame.
We compare to established methods and analyze the behavior of all algorithms along different experimental dimensions,
by varying the kinds and number of potential input features used,
number of shots chosen for training as well as the number of frames selected per shot.
We pay particular attention, besides the test accuracy, to train vs.\ test accuracy
(over-fitting) and training time.
We choose the large-scale YouTube-Objects video dataset~\cite{prest2012learning}, with
difficult sequences of ten categories (aeroplane, bird, boat, car, cat, cow, dog, horse, motorbike, train)
taken \emph{in the wild}. The training set contains about $4200$
video shots, for a total of $436970$ frames, while the test set has $1284$ video shots for a total of
over $134119$ frames.
The videos display significant background clutter, with objects coming in and out of
foreground focus, undergoing occlusions and significant changes in scale and viewpoint. More importantly,
the intra-class variation is large and sudden between video shots. Given the very large number of frames and variety of shots,
their complex appearance and variation in length, presence of background clutter and many other objects,
changes in scale, viewpoint and drastic intra-class variation, the task of recognizing the main category
from only a few frames becomes a real challenge. We used the same training/testing
split as in~\cite{prest2012learning}. In all our tests, we present results averaged over $30-100$ random experiments, for all methods compared.

\begin{table}
\caption{Distribution in percentages of sub-windows (Fig.~\ref{fig:diverse_features}) for selected ImageNet classifiers per category.  Note that different categories that seem superficially similar (e.g., cats and dogs) generate very different distributions (see text).}
\label{tab:prob_locations}
\begin{center}
\begin{tabular}{|l|r|r|r|r|r|r|}
\hline   Locations   & W       & C     & TL & TR & BL & BR  \\
\hline
\hline   aeroplane   &  65.6    &   30.2    &   0     &    0          & 2.1           &   2.1   \\
\hline   bird        &  78.1    &   21.9    &   0     &    0          & 0               &   0       \\
\hline   boat        &  45.8    &   21.6    &   0     &    0          & 12.3          &   20.2  \\
\hline   car         &  54.1    &   40.2    &   2.0 &    0          & 3.7           &   0       \\
\hline   cat         &  76.4    &   17.3    &   5.0 &         0     & 1.3           &   0       \\
\hline   cow         &  70.8    &   22.2    &   1.8 &    2.4      &       0         &   2.8   \\
\hline   dog         &  92.8    &    6.2    &   1.0 &         0     &   0             & 0         \\
\hline   horse       &  75.9    &   14.7    &       0 &         0     &   8.3         & 1.2     \\
\hline   motorbike   &  65.3    &   33.7    &       0 &         0     &   0             & 1.0     \\
\hline   train       &  56.5    &   20.0    &       0 &    2.4      &   12.8        & 8.4     \\
\hline
\end{tabular}
\end{center}
\end{table}

We evaluated six methods: ours, SVM on all input features, AdaBoost on all input features, ours with SVM
(applying SVM only to features selected by our method,
idea related to~\cite{nguyen2010optimal,weston2000feature,kira1992feature}),
forward-backward selection (FoBa)~\cite{zhang2009adaptive}
and simple averaging over all input features.
Recognition rate is computed per frame.
Input features have soft-values between $0$ and $1$ and
are expected to be positively correlated with the positive class (we remember during training which feature should be
flipped for which class). For our method, which outputs a sparse solution as a weighted
average over a few features, we \emph{select} those with a weight larger than a very small threshold.
Note that once features are selected, in principle, any classifier could be learned, to fine-tune
the weights, as is the case with \emph{ours with SVM}. While FoBa works directly with the features given,
AdaBoost further transforms each feature into a weak hard classifier by choosing the threshold
that minimizes the expected exponential loss, at each iteration; that is one reason why AdaBoost is much
slower w.r.t.\ to the others.

Table~\ref{tab:prob_locations} summarizes the locations distribution of ImageNet features selected by our method for each category in YouTube-Objects.  We make several observations.  First, the majority of features for all classes consider the whole image (W), which suggests that the image background is relevant. Second, for several categories (e.g., car, motorbike, aeroplane), the center (C) is important. Third, some categories (e.g., boat) may be located off-center or benefit from classifiers that focus on non-central regions.  Finally, we see that object categories that may superficially seem similar (cat vs.\ dog) exhibit rather different distributions: dogs seem to benefit from the whole image while cats benefit from sub-windows; this may be because cats are smaller and appear in more diverse contexts and locations, particularly in YouTube videos.
We evaluated the performance of all methods by varying the number of shots randomly chosen for training
and averaged the results over $30-100$ experiments.
\begin{figure}
\begin{center}
\includegraphics[scale = 0.47, angle = 0, viewport = 0 0 600 970, clip]{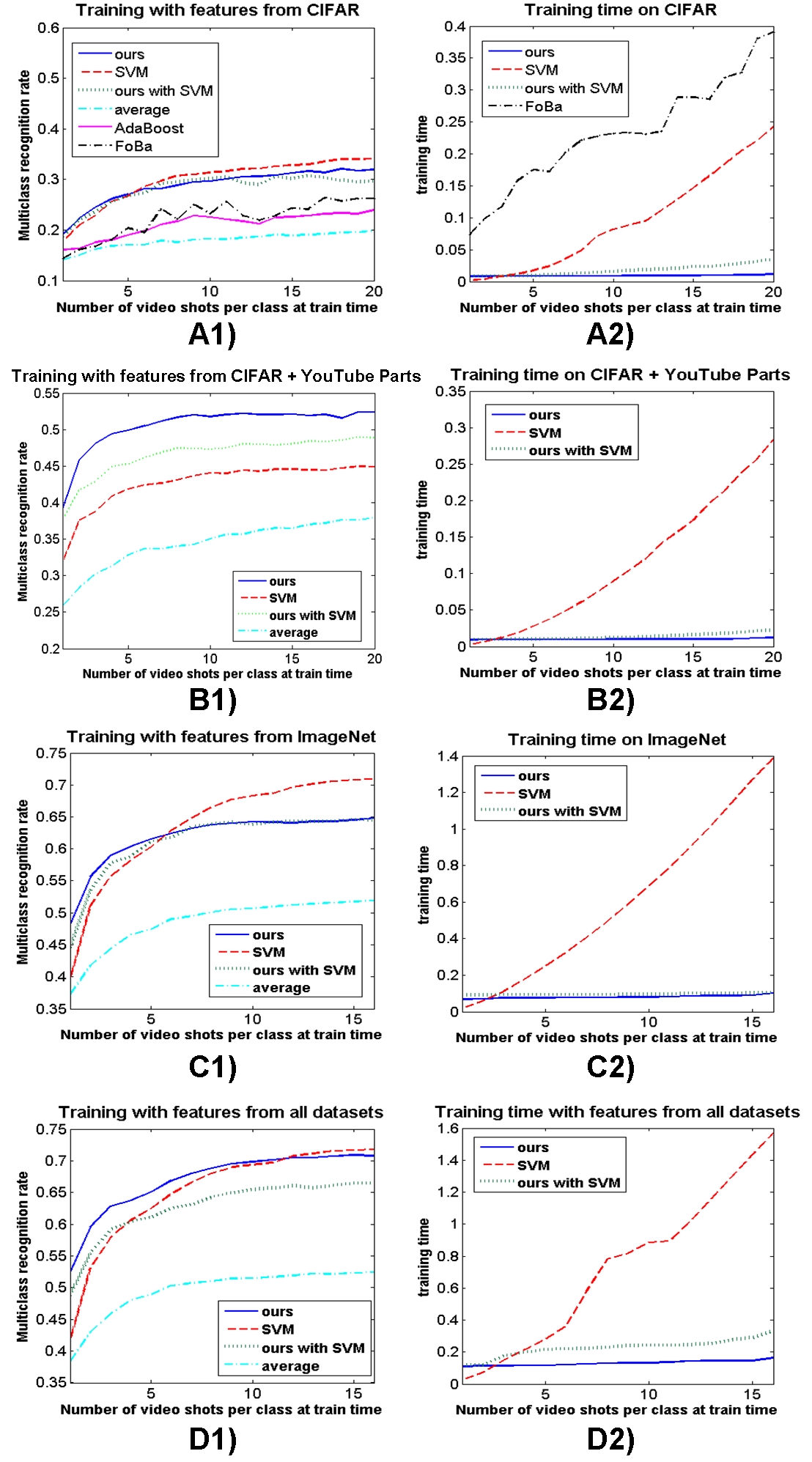}
\caption{Accuracy and training time on YouTube-Objects, with varying training video shots ($10$ frames per shot and results averaged over $30$ runs).
Input feature pool,
row~1: $50$ type~I features on CIFAR;
row~2: $110$ type~II features on YouTube-Parts + $50$ CIFAR;
row~3: $2000$ type~III features in ImageNet;
row~4: $2160$ all features.
Ours outperforms SVM, AdaBoost and FoBa (see text).
}
\label{fig:test_and_time}
\end{center}
\end{figure}

\begin{figure}
\begin{center}
\includegraphics[scale = 0.27, angle = 0, viewport = 0 0 900 530, clip]{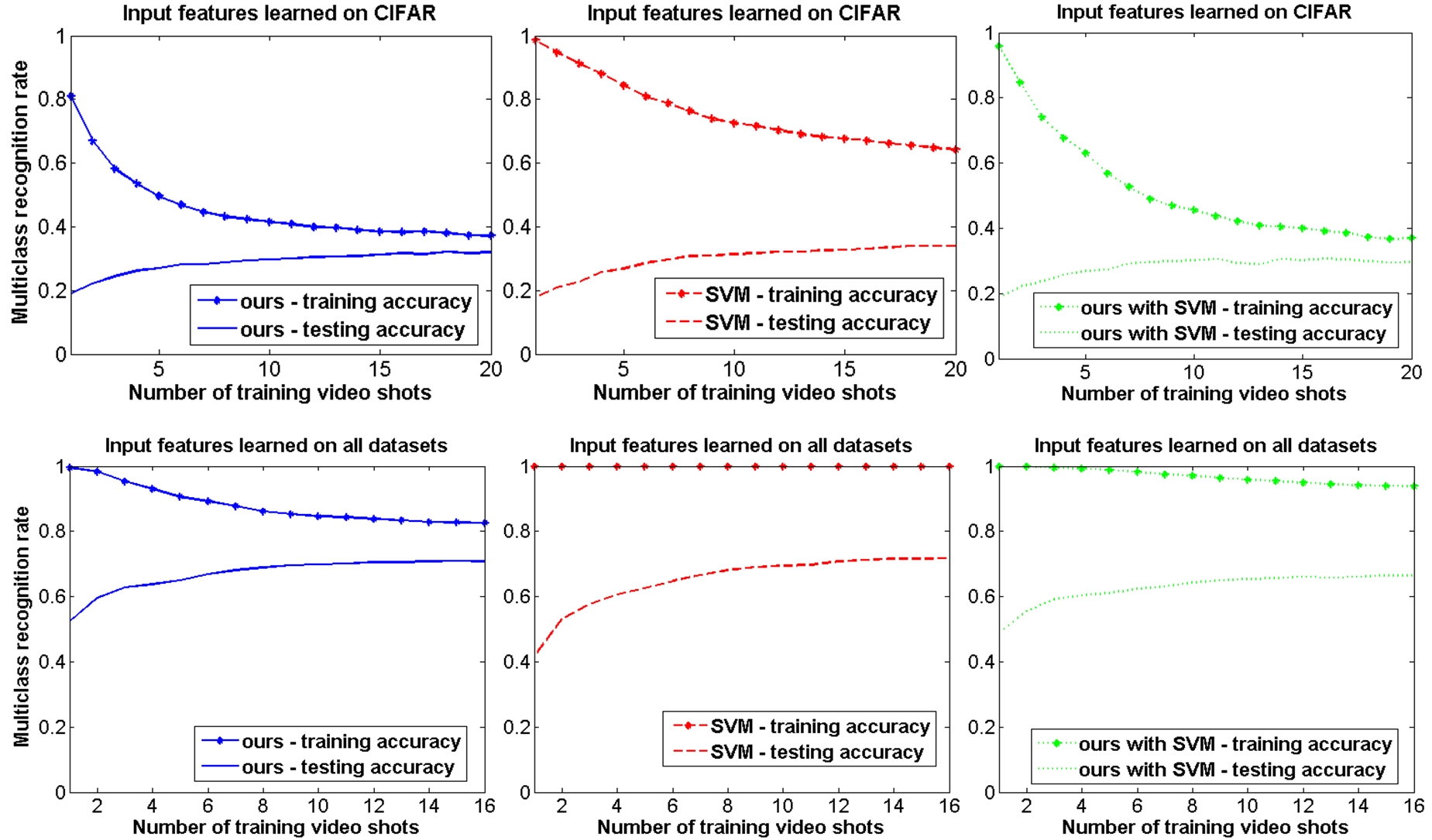}
\caption{Our method generalizes (training and test errors are closer) compared to SVM or in combination with SVM.}
\label{fig:train_vs_test}
\end{center}
\end{figure}

\begin{figure}
\begin{center}
\includegraphics[scale = 0.3, angle = 0, viewport = 0 0 700 290, clip]{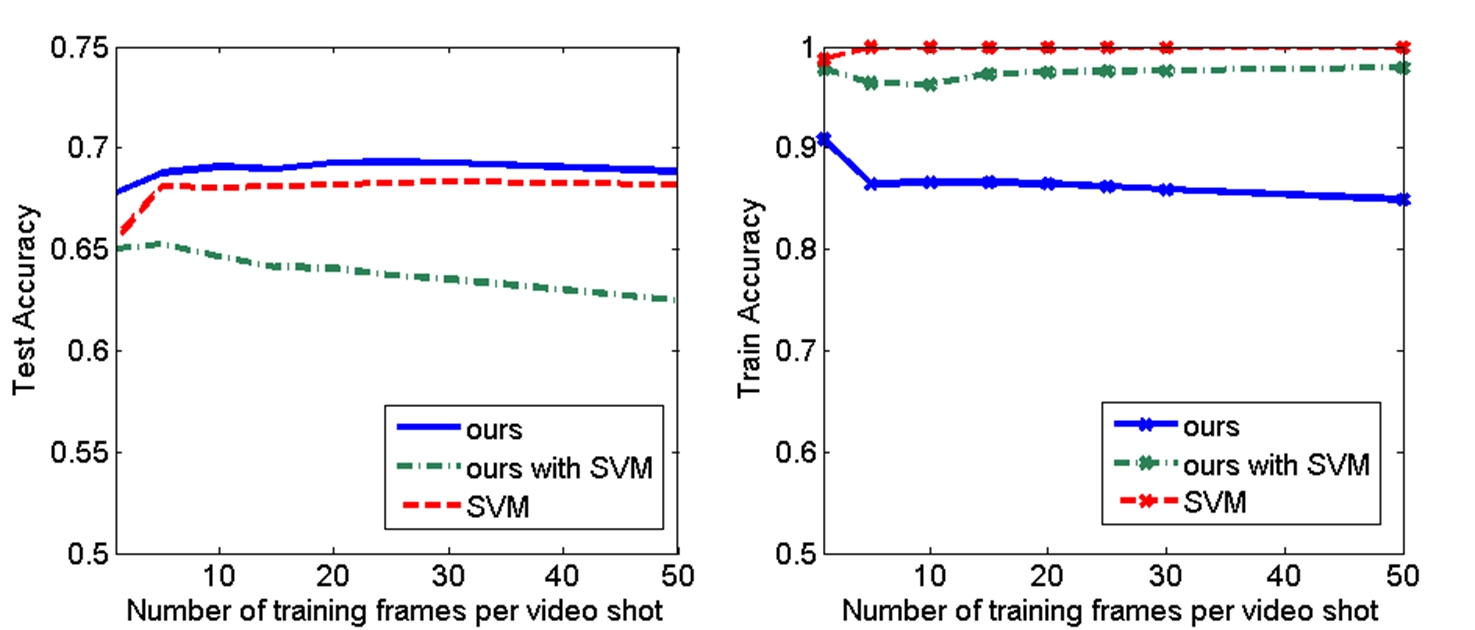}
\caption{Average test recognition accuracy over $30$ independent experiments of our method as we vary the number
of training frames uniformly sampled from random $8$ training video shots. Note how well our method generalizes from
as few as $1$ frame per video shot, for a total of $8$ positive training frames per class.}
\label{fig:accuracy_vs_nFrames}
\end{center}
\end{figure}

The results, presented in Fig.~\ref{fig:test_and_time},
show convincingly that our method has a constant training time, and is much less costly than
SVM, AdaBoost (time too large to show in the plot) and FoBa. Moreover, our method is able to outperform significantly most methods
(even SVM in many cases). As our intuition and theoretical results suggested, the proposed discriminative feature clustering
approach is superior to the others as the amount of training data is more limited
(also see Figs.~\ref{fig:train_vs_test} and~\ref{fig:accuracy_vs_nFrames}).
Our mining of powerful groups of classifiers from a vast sea of candidates from limited data
is a novel direction, complementary to learning approaches that spend significant training time and
data to fit optimal real weights over many features.
We also validate the importance of the feature pool size and quality
(Table~\ref{tab:testing_nFeatures}).

\begin{figure*}
\begin{center}
\includegraphics[scale = 0.35, angle = 0, viewport = 0 0 1400 1550, clip]{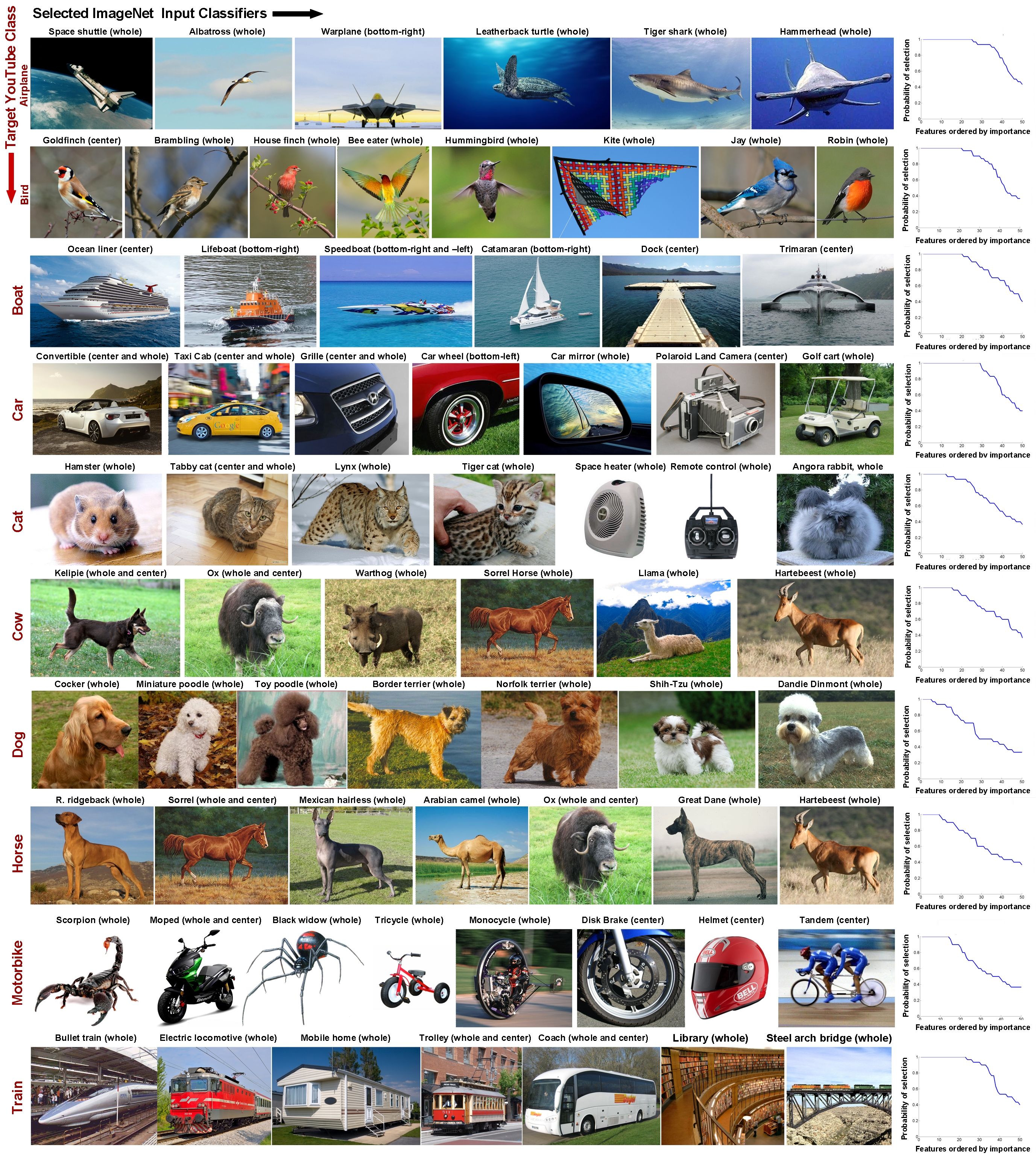}
\caption{For each training target class from YouTube-Objects videos (labels on the left), we present the most frequently selected
ImageNet classifiers (input features), over $30$ independent experiments, with $10$ frames per shot and
$10$ random shots for training. In images we show the classes that were always selected by our method when $k=50$.
On the right we show the probability
of selection for the most important $50$ features.
Note how stable the selection process is. Also note the interesting connections between the selected
classes and the target class in terms of appearance, context or geometric part-whole relationships.
We find two aspects indeed interesting: 1) the high probability (perfect 1) of selection
of the same classes, even for such small random training sets and 2) the fact that unrelated classes in terms of meaning
could be so useful for classification, based on their shape and appearance similarity.}
\label{fig:discovered_classes}
\end{center}
\end{figure*}

\begin{table}
\caption{Accuracy on YouTube-Objects with varying number
of training shots for different feature pools. Accuracy doubles with the size and diversity of the pool.
}
\label{tab:testing_nFeatures}
\begin{center}
\begin{tabular}{|l|r|r|r|}
\hline   Accuracy                 & I ($50$)       & I+II ($160$)    & I+II+III ($6160$) \\
\hline
\hline   $10$ train shots        &  29.69\%           &  51.57\%               &  69.99 \%                      \\
\hline   $20$ train shots        &  31.97\%           &  52.37\%               &  71.31 \%                      \\
\hline
\end{tabular}
\end{center}
\end{table}

\paragraph{Intuition and qualitative results:} An interesting finding in our
experiments (see Fig.~\ref{fig:discovered_classes})
is the consistent discovery, for a given target class, of selected
input classifiers that are related to the main one in surprising ways:
\begin{inparaenum}[1)]
\item similar w.r.t.\
global visual appearance, but not semantic meaning -- banister vs.\ train, tigershark vs.\ plane, Polaroid camera vs.\
car, scorpion vs.\ motorbike, remote control vs.\ cat's face, space heater vs.\ cat's head;
\item related in co-occurrence and context, but not in global appearance -- helmet vs.\ motorbike;
\item connected through part-to-whole relationships --
grille, mirror and wheel vs.\ car;
or combinations of the above -- dock vs.\ boat, steel bridge vs.\ train,
albatross vs.\ plane.
\end{inparaenum}
The relationships between the target class and the input, supporting classes,
could also hide combinations of many other factors.
Meaningful conceptual relationships could ultimately
join together correlations along many dimensions, from appearance to
geometric, temporal and interaction-like relations.

Another interesting aspect is that the classes
found are not necessarily central to the main category, but often peripheral, acting as guardians that separate
the main class from the rest. This is where feature diversity plays an important role, ensuring both \emph{separation}
from nearby classes as well as robustness to missing values.

An additional possible benefit is the capacity to
immediately learn novel concepts from old ones, by combining existing high-level concepts to recognize new classes. In cases where there is insufficient
data for a particular new class, sparse averages of reliable
classifiers can be an excellent way to combine previous knowledge.
Consider the class \emph{cow} in Fig.~\ref{fig:discovered_classes}.
Although ``cow'' is not present in the $1000$ label set, our method is
able to learn the concept by combining existing classifiers.

Since categories share shapes, parts and designs, it is perhaps unsurprising
that classifiers trained on semantically distant classes that are
visually similar can help improve learning and generalization from limited
data.


\section{Conclusions}

We have presented an efficient method for joint selection of discriminative
and diverse groups of features that are independent by themselves and strong in combination.
Our feature selection solution comes directly from a supervised linear classification problem
with specific affine and size constraints, which can be solved rapidly due to its convexity.
Our approach is able to quickly learn from limited data effective classifiers that outperform
in time and even accuracy more established methods such as SVM, Adaboost and greedy sequential selection.
We also propose different ways of creating novel, diverse features, by learning separate classifiers
over the input space and over different regions in the input image.
Having a training time that is independent of the number of input images and an effective way of learning
from large and heterogeneous feature pools, our approach provides a useful tool for many recognition
tasks, suited for real-time, dynamic environments. Based on our extensive experiments
we believe that it has the potential to strengthen the connection between the apparently separate
problems of unsupervised clustering, linear discriminant analysis and feature selection.

\paragraph{Acknowledgments:}
This work was supported by CNCS-UEFICSDI, under project PNII PCE-2012-4-0581.
The authors would like to thank Shumeet Baluja for interesting discussions and
helpful feedback.
%
%
%

\FloatBarrier	
\bibliographystyle{ieee}
\bibliography{complete}

\end{document}